\documentclass[letterpaper, 10 pt, conference]{ieeeconf}  

\IEEEoverridecommandlockouts                              

\usepackage{bm, amsmath} 
\usepackage{cclicenses, graphicx}
\usepackage{cite}

\newcommand{\mylist}{\begin{list}{$\bullet$}
    {\leftmargin6mm \itemsep0pt \itemindent0mm \topsep1mm \parsep1mm \labelsep4mm}
}

\newcommand{\mylisttight}{\begin{list}{$\bullet$}
    {\leftmargin0mm \itemsep0pt \itemindent0mm \topsep0mm \parsep0.5mm}
}

\title{\LARGE \bf
Proprioception-Based Grasping for Unknown Objects\\ Using a Series-Elastic-Actuated Gripper*
}

\author{Tianjian Chen$^{1}$ and Matei Ciocarlie$^{1}$
\thanks{*This work was supported in part by the NASA Early Space Innovations program through award NNX16AD13G.}
\thanks{$^{1}$Both authors are with the department of Mechanical Engineering, Columbia University, New York, NY 10027, USA.
        {\tt\small \{tc2764, matei.ciocarlie\}@columbia.edu}}%
}

\begin{document}

\maketitle
\thispagestyle{empty}
\pagestyle{empty}

\begin{abstract}
Grasping unknown objects has been an active research topic for
decades. Approaches range from using various sensors  (e.g. vision, tactile) to gain
information about the object, to building
passively compliant hands that react appropriately to contacts. In this
paper, we focus on grasping unknown objects using proprioception (the
combination of joint position and torque sensing). Our hypothesis is
that proprioception alone can be the basis for versatile
performance, including multiple types of grasps for objects with multiple
shapes and sizes, and transitions between grasps. Using a series-elastic-actuated gripper, we propose a method for performing stable fingertip
grasps for unknown objects with unknown contacts, formulated as multi-input-multi-output
(MIMO) control. We
also show that the proprioceptive gripper can perform enveloping
grasps, as well as the transition from fingertip grasps to enveloping
grasps.

\end{abstract}

\section{Introduction}

Proprioception, or the ability to perceive the relative positioning of
neighboring body parts as well as the muscle effort deployed to
produce it, is a fundamental human sense. Together with vision and
tactile sensing, it plays a unique and important role in human hand
perception \cite{blanchard2013differential}. For robotic manipulation,
proprioception is translated as the combination of joint position and
torque sensing (assuming a hand comprised exclusively of revolute joints).

Compared to vision and tactile sensing, both of which have been
studied extensively in the context of manipulation, we believe that
grasping with proprioception is still an
important area to advance. On one hand, vision and tactile sensing
have intrinsic limitations, such as occlusion for vision and hardware
complexity for tactile sensing. On the other hand, when all these
senses are available, they can still complement each other.  Demonstrating
manipulation capabilities based exclusively on proprioception becomes
a useful exercise: we believe the more a hand can do with only one
sensing modality, the more versatile it will be when multi-modal
sensory information gets integrated.

As we show here, proprioception is promising in providing the hand
with the ability to adapt to the previously unknown shape of the
object, and to execute stable grasps. We note that there are multiple
ways for hands to adapt to an object: while fully-actuated
hands use sensor information (such as proprioception here) to perform
active adaptation, underactuated hands are good examples of passive
adaptation without the use of sensing. However, the former have the
advantage of versatility: proprioceptive grippers can provide
compliance similar to passive mechanisms, but can also change behavior
at runtime and selectively execute different types of grasps. Of
course, the price paid for the additional versatility is the increased
complexity of the sensory setup.

In this study, we explore the problem of \textit{grasping using only
proprioceptive feedback}, without any contact information or
knowledge of object pose and properties. To the best of our knowledge,
we are the first to show that a robot hand can perform all the
following tasks using proprioception exclusively:
\begin{itemize}
\item execution of fingertip grasps for unknown objects;
\item execution of enveloping grasps for unknown objects;
\item on-demand transitions between fingertip and enveloping grasps.
\end{itemize}
Our main contributions are to provide methods for the tasks above,
and, in the process, demonstrate their effectiveness by
experiment. Our results indicate that the proprioceptive gripper is
more versatile in the range of fingertip grasps it can perform,
compared to our two baselines: an emulated underactuated gripper
commanding fixed torques to the joints, as well as a physically
constructed underactuated gripper. In addition, our gripper also
displays the ability to execute enveloping grasps and to transition to
them from fingertip grasps. Both examples of increased versatility
were achieved using proprioception as the only available sensing
modality.

\section{Related Work}

Researchers have been exploring real-time sensing and control as
an alternative to vision-based planning in manipulation. Assuming
object information is available, model-based controllers can perform
grasping or in-hand manipulation. For example, Yoshikawa et
al. presented studies (e.g. \cite{yoshikawa2000control}) on hybrid
force-position control for manipulation.  Arimoto et
al. \cite{arimoto2000dynamics} derived the dynamics of a dual-finger
gripper and proposed a controller which can regulate the object
position and orientation. Caccavale et al. \cite{caccavale2013grasp}
proposed an impedance controller to keep track of desired object
trajectory and ensure the grasp quality simultaneously. Unlike our
approach, these methods require complete information of the
hand-object system.

When the models of the objects are not available, researchers either
relied on assumption about the contacts, or used sensor-based
techniques for grasping. For example, Schneider and
Cannon \cite{schneider1992object} studied object impedance control
using multiple manipulators. Arimoto et al. \cite{arimoto2005two} and Yoshida et al. \cite{yoshida2007blind}
studied ``blind grasping'' using two fingertips. However, these studies
assume the contacts only happen at the end points or the end hemispheres
of the fingertips. Wang et al. \cite{wang2007switching} proposed a
controller that can search appropriate finger contact locations using
haptic feedback and can switch between control modes for different
surfaces. Platt et al. \cite{platt2010null} presented a study on
changing the contact configuration by following the gradient of
grasping objective functions using six-axis loadcell data. Hsiao et
al. \cite{hsiao2010contact} proposed a contact-reactive method using
tactile sensing to deal with uncertainty. However, these methods
require contact sensing methods, such as tactile sensors or in-finger
load cells. In contrast, our approach does
  not make any assumptions about contact location or state, and does not
require tactile sensing data.

Torque measurement is often used for grasp control. Researchers have
developed several robotic hands with force or torque sensing. For
example, the Robonaut Hand \cite{lovchik1999robonaut} and the DLR Hand
II \cite{butterfass2001dlr} have strain gauges or force-torque sensors
embedded in their fingers. The hand of the DOMO robot
\cite{edsinger2004domo} and the hand of the Obrero robotic platform
\cite{torres2005obrero}
make use of the Series Elastic Actuators (SEA), which are a type of
actuators with elastic components in series with the motor to sense
the torque \cite{pratt1995series}. Furthermore, the DLR Hand-Arm
System \cite{grebenstein2011dlr} incorporates the Variable Impedance
Actuators, which are SEAs whose spring stiffnesses are actively
controlled. These hardware designs offer high performance, but at the
cost of high complexity and large overall packages.

As an alternative, researchers have developed underactuated hands that do
not require sophisticated sensing and control, and this types of hands are good baselines to compare against. Underactuated hands can
adapt to the object and make a grasp by the virtue of
carefully-designed torque ratios between joints. The Harvard Hand
\cite{dollar2010highly}, iHY Hand \cite{odhner2014compliant}, Robotiq
Hand \cite{birglen2004kinetostatic}, and Velo Gripper
\cite{ciocarlie2014velo} are good examples in this
category. However, even though underactuation simplifies control, it
generally does not provide as much dexterity as full actuation. Many
of the hands above can only perform certain types of grasps, or lack the flexibility to choose the
configuration after making the grasp.

\begin{figure}[t]
\centering
\includegraphics[width = 67 mm]{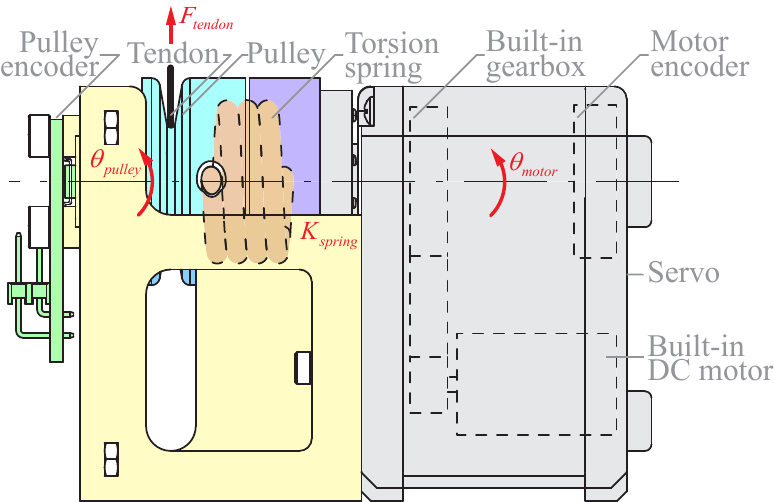}
\caption{SEA module.}
\label{fig:sea}
\vspace{-4mm}
\end{figure}

\section{Hardware Platform}

While joint position sensing is ubiquitous for fully-actuated robot
hands, torque sensing and control is not common in commercially available
manipulators. This compelled us to design our own hardware testbed. We
implemented torque sensing and control with Series Elastic Actuators
(SEA), a method known for high-fidelity torque control, shock
protection, and human-safety~\cite{pratt1995series}. 

\subsubsection{SEA Module}

Similar to the design from Ates et al. \cite{ates2014servosea}, we developed a simple and compact SEA module (Fig.~\ref{fig:sea}). A position-driven servo (gray) is used as the driving motor, which receives position commands and returns the current position (measured by the built-in potentiometer), i.e., $\theta_{motor}$  can be measured. A torsion spring (orange) is used as the elastic component, connecting the motor shaft (purple) and the pulley shaft (blue). A Spectra cable is tied on the pulley to transmit the force to the finger joint. An absolute magnetic encoder (in green) is mounted on the end of the pulley shaft to measure $\theta_{pulley}$. In steady-state, the force in the tendon can be calculated as the product of spring stiffness and deflection divided by pulley radius: 
\begin{equation} \label{eq:sea}
F_{tendon} = K_{spring} \cdot (\theta_{pulley} - \theta_{motor}) / R_{pulley} 
\end{equation}

\begin{figure}[t]
\centering
\includegraphics[width = 80 mm]{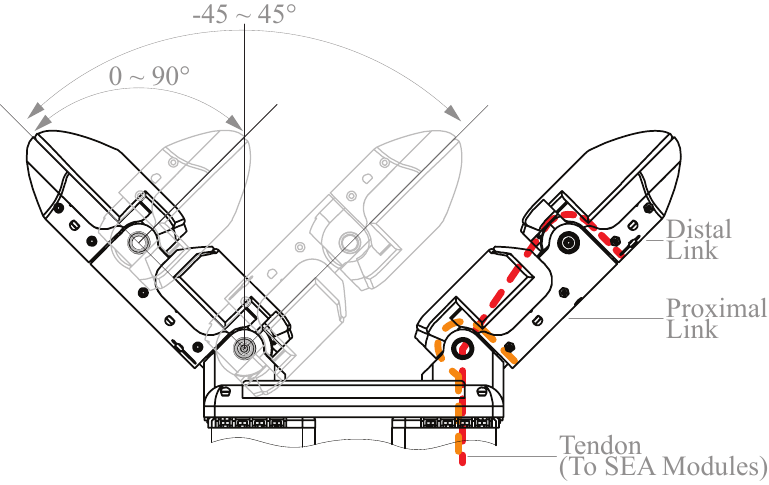}\\
\caption{Schematics of the gripper.}
\label{fig:hand}
\vspace{-4mm}
\end{figure}

\subsubsection{Gripper Design}

The gripper consists of two fingers and each finger has two links, shown in Fig.~\ref{fig:hand}. 
In each joint, flexion is
powered by the tendon (shown as colored dash lines in
Fig.~\ref{fig:hand}) connected to the SEA pulley, and the extension is
driven by a restoring spring. The tendon connected to the
distal joint (the red dash line) goes right through the axis of the
proximal joint so that the torques of proximal and distal joints are
fully decoupled.


\subsubsection{SEA-level Control}

There are three SEA-level control modes: motor position control, pulley position control and torque control, and they can be switched online. 
We note that we built joint-level or hand-level controllers on top of these SEA-level controllers. For example, joint position and torque control can be achieved by SEA pulley position and torque control with a simple linear conversion.


\begin{figure*}[t]
\centering
\includegraphics[width = \textwidth ]{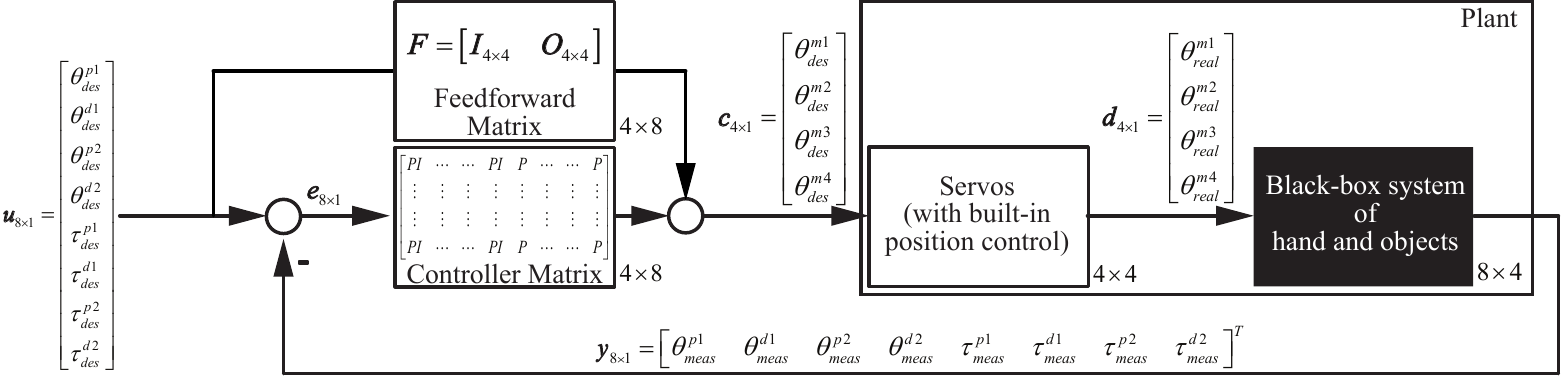}
\caption{Schematics of the MIMO Grasping Controller. (Subscripts: des---desired value, meas---measured value, real---real value. Superscripts: m---motor, p---proximal link, d---distal link).}
\label{fig:mimo}
\vspace{-4mm}
\end{figure*}

\begin{figure}[t]
\centering
\includegraphics[width = 80mm ]{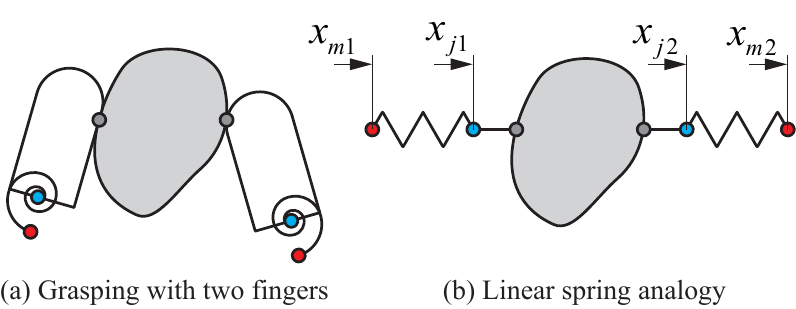}
\caption{Problem illustration.}
\label{fig:analogy}
\vspace{-4mm}
\end{figure}

\section{Fingertip Grasping}

Fingertip grasps commonly refer to grasps where only the most
distal links of each finger make contact with the object. We note that it does not necessarily mean the contacts are located in the very end of the fingers, so contact locations are still unknown. This type of
grasp is important not only for precision tasks, but also for cases
where a more stable enveloping grasp is not immediately available
because of the environment (an object laying on a table, against a
backdrop, etc.). In this section, we introduce a control algorithm
which can perform stable fingertip grasps for unknown objects.

\subsection{Problem Statement}

The objective of our algorithm is to
increase the torques applied at the joints (and implicitly the contact forces) in a ``stable'' fashion after making initial contacts. In other words, we need to find an
increase in joint torques that produces no net wrench on the
object. For a hand with multi-link fingers, this is not straightforward given that the joint torques need to be coordinated in the absence of information on object
shape and contact locations. It is necessary to note that we are not
solving contact planning problem, and we wish to develop reactive control strategies that do not require pre-planning.

Our insight is that proprioception alone can characterize grasp stability. For an SEA-powered proprioceptive gripper, an
unbalanced net wrench will produce object movement against the
compliant elements, which we can measure. Therefore, \textit{we
  formulate the goal of grasp stability as the one of minimizing
  object movement while applying forces}. Since we do not have a direct measure of object
movement in Cartesian coordinates by using only proprioception, we use the change of joint position in joint space (measured by SEA)
during grasping as a proxy for object movement.

For intuition, consider the analogies shown in Fig.~\ref{fig:analogy}:
Figure (a) shows a (simplified) scenario in which the motors (red dots) are
driving the torsion springs, and the springs are pushing the fingers
(blue dots) to make a grasp (gray dots). (b) shows a simpler
one-dimensional abstraction using linear springs, with similar
color-coding as in (a). Here, we actively control the positions
$x_{m1}$ and $x_{m2}$ (which translate to motor positions
$\theta_{motor}$ on the real gripper) to apply forces, and measure
$x_{j1}$ and $x_{j2}$ (which translate to pulley positions
$\theta_{pulley}$ linearly mapped to joint positions). We aim to keep
$x_{j1}$ and $x_{j2}$ constant as we squeeze.

\subsection{MIMO Grasping Controller}

Our key insight is that the problem of grasping unknown objects can be solved even in the absence of
contact information, by using proprioception as inputs for a
multi-input-multi-output (MIMO) proportional-integral (PI)
control. Without knowledge of object geometry, contact locations and contact states, it
is impossible to fully model the dynamic system analytically. However,
a proprioceptive platform still provides sensory access to the
variables that characterize the grasp stability, and the PI control framework provides ways to regulate these variables even though the analytical relationship is not constructed . We
thus aim to use a feedback scheme operating exclusively in
the sensory space of the robot, without explicitly modeling the physics of the gripper and the object.

Fig.~\ref{fig:mimo} shows the block diagram of the MIMO control loop. Here, the controller is constructed on top
of the low-level sensing, so we consider the joint angles and torques
are already obtained from SEA measurements. The reference vector
$\bm{u}$ consists of desired joint angle values ($\theta^{p1}_{des}$,
$\theta^{d1}_{des}$, $\theta^{p2}_{des}$, $\theta^{d2}_{des}$, where
the superscripts $p$ represent proximal and $d$ represent distal
joints) and reference joint torques ($\tau^{p1}_{des}$,
$\tau^{d1}_{des}$, $\tau^{p2}_{des}$, $\tau^{d2}_{des}$ ). The feedback vector $\bm{y}$ has the same structure, but contains actual measured
measured values. The desired joint angles (first half of the reference
vector $\bm{u}$) are extracted by a feedforward matrix $\bm{F}$ and
used as a feedforward term. The error between the reference $\bm{u}$
and feedback $\bm{y}$ is fed into a MIMO PI block (a combination of
many P and PI controllers) which is a $4\times8$ matrix. The output of
the PI block and the feedforward term are summed up as motor position
command $\bm{c}$ (a $ 4 \times 1$ vector) and sent to the motors:
\begin{equation} \label{eq:mimo}
\bm{c}(t)=\bm{F u}(t)+ {\bm{K}_{p}}  \bm{e}(t)+ {\bm{K}_{i}} \int_{0}^{t}{ \bm{e}(\tau) d\tau} 
\end{equation}
Here,
$
\bm{F}=\left[ \begin{matrix} {\bm{I}_{4\times 4}} & {\bm{O}_{4\times 4}}\end{matrix} \right]
$
is the feedforward matrix,
$\bm{K}_p$ ($4\times 8$ matrix with all entries being non-zero) and 
$\bm{K}_i=\left[ \begin{matrix} {\bm{K}_{4\times 4}} & {\bm{O}_{4\times 4}}\end{matrix} \right]$ (where $\bm{K}$ is a matrix with all entries being non-zero)
are proportional and integral gain matrices, and $\bm{e}$ is the $ 8 \times 1$ error vector between $\bm{y}$ and $\bm{u}$. After that, the actual motor position vector $\bm{d}$ goes to the black-box system of the gripper and the unknown object.

In the reference vector $\bm{u}$, the desired joint angle values  ($\theta^{p1}_{des}$, $\theta^{d1}_{des}$, $\theta^{p2}_{des}$, $\theta^{d2}_{des}$) 
are equal to those in the initial touch configuration, while the reference torques ($\tau^{p1}_{des}$, $\tau^{d1}_{des}$, $\tau^{p2}_{des}$, $\tau^{d2}_{des}$ ) 
are chosen using the maximum motor torques. A special design of this controller is that we require the joint angles to be regulated exactly to the set points, but do not require the torques to be so. We allow and make use of the steady-state error of pure proportional control (we note that entries in the right half of the integral gain $\bm{K}_i$ are set to be zeros). In this way, the reference torques do not need to be a legal set of torques that result in equilibrium --- actually, we are not able to design such a legal set of torques due to the absence of contact or object information. We let the law of dynamics decide the steady-state values for torques, and let the system balance itself automatically. The effectiveness of increasing joint torques is shown in section VI.

From a practical standpoint, the tuning process of the MIMO PI controller is not
as complicated as it would seem based on the number of parameters. First, due to gripper symmetry, the number of parameters is cut by half. Second, we formulate
every gain as a product of a baseline value ($b_i$ in (\ref{eq:kp})(\ref{eq:ki}) )  and a weight
coefficient ($w_i$ in (\ref{eq:kp})(\ref{eq:ki}) ). The baseline values are set to be the same if the input
entries corresponding to those gains have same physical
dimensionality, and the weight coefficients are tuned based on its
relative importance. Third, conventional tuning heuristics for the gains of single-input-single-output systems also apply here. The structures of the gain matrices are as follows:
\begin{equation} \label{eq:kp}
\bm{K}_p=\left[ \begin{smallmatrix}
{w_1 b_1} & {w_2 b_2}  & {w_2 b_1}  &  {w_2 b_2}  & {w_3 b_3}  & {w_4 b_4} & {w_4 b_3} & {w_4 b_4} \\
{w_2 b_1} & {w_1 b_2}  & {w_2 b_1}  &  {w_2 b_2}  & {w_4 b_3}  & {w_3 b_4} & {w_4 b_3} & {w_4 b_4} \\
{w_2 b_1} & {w_2 b_2}  & {w_1 b_1}  &  {w_2 b_2}  & {w_4 b_3}  & {w_4 b_4} & {w_3 b_3} & {w_4 b_4} \\
{w_2 b_1} & {w_2 b_2}  & {w_2 b_1}  &  {w_1 b_2}  & {w_4 b_3}  & {w_4 b_4} & {w_4 b_3} & {w_3 b_4} \\
\end{smallmatrix} \right]
\end{equation}
\begin{equation} \label{eq:ki}
\bm{K}_i=\left[ \begin{smallmatrix}
{w_1 b_5} & {w_2 b_6}  & {w_2 b_5}  &  {w_2 b_6}  & ~~0~~ & ~~0~~ & ~~0~~ & ~~0~~ \\
{w_2 b_5} & {w_1 b_6}  & {w_2 b_5}  &  {w_2 b_6}  &   0 & 0  & 0  &     0\\
{w_2 b_5} & {w_2 b_6}  & {w_1 b_5}  &  {w_2 b_6}  &   0  & 0  & 0  &    0 \\
{w_2 b_5} & {w_2 b_6}  & {w_2 b_5}  &  {w_1 b_6}  & 0 & 0 & 0 & 0 \\
\end{smallmatrix} \right]
\end{equation}
We pick $b_1 = 0.2$, $b_2 = 0.5$, $b_3 = 4.0$, $b_4 = 8.0$, $b_5 = 1.0$, $b_6 = 1.0$, $w_1 = 1.0$, $w_2 = 0.3$, $w_3 = 1.0$, $w_4 = 0.5$ for our hardware.

\section{Enveloping Grasping and Transitions}

An enveloping grasp is the one where both distal and proximal links
make contact with the object around its circumference. This type of
grasp is generally considered more stable than a fingertip grasp
because it can resist disturbances in a wider range of directions. The
inability to envelop is also one of the main shortcomings of simple
parallel grippers. In contrast, some of the more recent underactuated
hands are optimized explicitly for effective enveloping grasps of a
wide range of objects (e.g. \cite{ciocarlie2014velo}).

When using our proprioceptive gripper, we found that stable
enveloping grasps for unknown objects are easier to obtain than fingertip grasps. The
mechanism is generally fully constrained and all the links are
counterbalancing each other. A simple joint torque control scheme, or the MIMO Grasping Controller, can fulfill this task.

A very important ability of this gripper, further underlining its
versatility, is to \textit{transition} between grasp types when holding unknown objects. After
executing a stable fingertip grasp (using the MIMO Grasping
Controller), the gripper can switch to joint torque control with the
torque ratio (between distal and proximal joints) being 0.5 to 1.0,
thus bringing the object into the hand and creating an enveloping
grasp. We illustrate this behavior with several experiments in the
following section.

\section{Experiments and Results}
In this section, we demonstrate the merits of the
proprioception-enabled gripper by several experiments. We validated
its capability of performing fingertip grasps, enveloping grasps, and
the transition from the former to the latter. We note that in this
study we only consider a two-dimensional scenario, in which the
objects are confined to move only in the plane of the fingers.

\subsection{Fingertip Grasp}
The goal of this experiment is to test the hypothesis that the MIMO Grasping Controller is effective in fingertip grasping for unknown objects. We compare against two baselines: a (fully-actuated) gripper running a Fixed Torque Ratio Controller, and a physical underactuated gripper. 

The Fixed Torque Ratio Controller, where the torques applied to proximal and distal joints always follow a certain ratio, can be thought of as an emulation of a common type of underactuated grippers (tendon-pulley-driven, without special designs such as stoppers or clutches). When this kind of grippers make grasps, the configuration-dependent torques from the extension springs can be ignored, as they are usually much smaller than the flexing torques from the tendons. Therefore, the net joint torques in proximal and distal joint have a configuration-independent and design-time-fixed ratio, which is the ratio between the joint pulley radii.

Furthermore, we understand that this emulation is subject to limited control bandwidth and may have unrealistic behavior compared to its physical counterpart. Thus, we also built a physically underactuated gripper testbed for comparison. This testbed has same specs as the fully-actuated proprioceptive gripper, except that the proximal and distal joints are driven by a single tendon wrapping around the joint pulleys. In this design, we can alter the torque ratio by physically changing the pulleys between experiments. We perform torque control for proximal joints in our experiments, thus the torques on the distal joints are defined by the physically determined ratios. However, in this setup, we lose the ability to measure joint positions by SEA readings because, in underactuated mechanisms, joint positions are determined not only by actuator positions but also by contact forces which here are unknown.

\subsubsection{Experiment Protocol}

Our experiment proceeds as follows. We execute the grasping in two phases. In the first phase (approaching and touching), the fingers are set in torque control mode with very low reference torques so that they stop when they touch the object. In the second phase(squeezing), the gripper executes the MIMO Grasping Controller or the Fixed Torque Ratio Controller in the fully-actuated testbed, or the  joint torque control on proximal joints in the underactuated testbed for comparison.

\begin{figure}[t]
\centering
\includegraphics[width=55 mm]{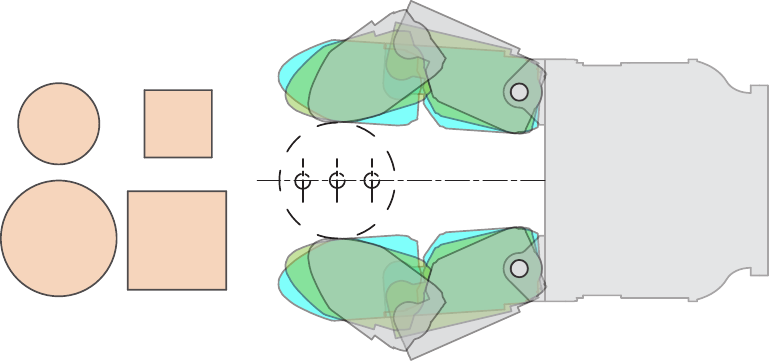}\\
\caption{Object sizes, object locations and initial touch poses.}
\label{fig:expsetup}
\vspace{-3mm}
\end{figure}

\begin{figure}[t]
\hspace{-6mm}\includegraphics[width=100 mm]{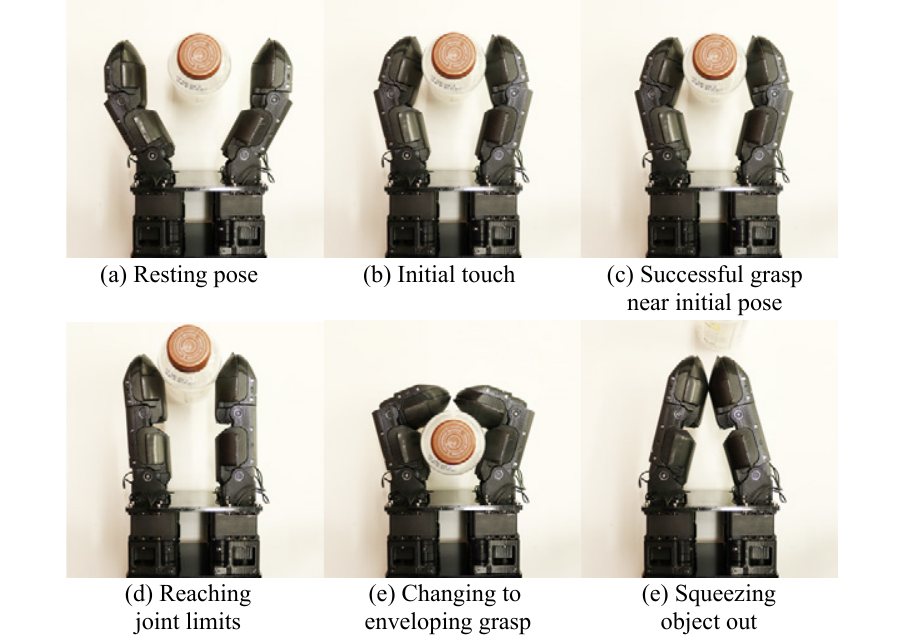}
\caption{Pose changes in different controllers.}
\label{fig:pose}
\vspace{-3mm}
\end{figure}

There are a lot of factors that may influence the performance. To have a well-rounded comparison, we swept the following dimensions:
\begin{itemize}
\item \textit{Controllers.} The torque ratio is a key parameter for both the Fixed Torque Ratio Controller, and the physically underactuated gripper. We tested the MIMO Grasping Controller against the other two baselines with three different ratios between the distal and proximal joint: 0.3, 0.4 and 0.5.
\item \textit{Objects.} We selected four objects for the test: a big cylinder (diameter: 67mm), a big box (side length: 57mm), a small cylinder (diameter: 47mm) and a small box (side length: 39 mm). All objects have negligible friction with the table.
\item \textit{Object locations.} We swept three locations along the center line of the gripper within the range of fingertip grasp: 100 mm, 120 mm and 140 mm from the palm.
\item \textit{Initial touch poses.} We tested three different distal joint angles for the initial touch: 0, 30 and 60 degrees. We note that we only include this dimension for MIMO Grasping Controller and Fixed Torque Ratio Controller test, and not for the physically underactuated hand because the distal joint angles of initial touch cannot be explicitly controlled in runtime.
\item \textit{Friction coefficient.} We tested the controllers with
  two fingertip materials: rubber (high friction, $\mu = 1.2$) and
  vinyl plastic (low friction, $\mu = 0.4$).
\end{itemize}

To sum up, we swept all five dimensions and conducted 360 grasping experiments. Fig.~\ref{fig:expsetup} shows the three object locations (shown as the crosshairs), three initial touch poses (colored fingers), and the sizes of the objects relative to the gripper (orange shapes).

\begin{figure}[t]
\centering
\includegraphics{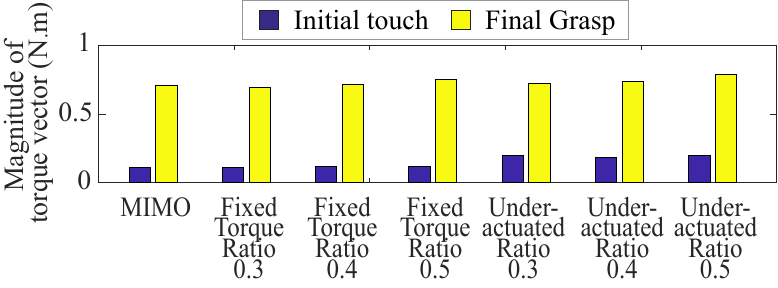}
\caption{Increases of torque magnitude during squeezing}
\label{fig:trq}
\vspace{-3mm}
\end{figure}

\subsubsection{Performance Metric}

\begin{itemize}
\item \textit{Success rate.} The success rate is our primary performance metric. We define a ``success" if the gripper finally settles down in equilibrium with the object in hand after squeezing. This definition includes three scenarios: (1) the gripper keeps the object in fingertips near initial touch pose, without converting to enveloping grasp or reaching joint limits, (2) the gripper holds the object but reconfigures to an enveloping grasp, and (3) the gripper keeps the object in fingertips but reaches a mechanical joint limit (thus the joint torque ratio changes). These cases are all considered successful but still need to be distinguished. Case (1) is the most desirable, while (2) and (3) mean the grasp is not stable at initial pose and relies on reconfiguration to be balanced.

\item \textit{Gripper pose change.} We believe it is also useful to keep the object in the same pose as when first contact is made. We thus use a secondary performance metric that evaluates how much the object moves in the hand during the squeezing process, with less movement considered better. Without access to object pose in Cartesian space, we measure this as the change in gripper pose between initial touch and final grasp (Euclidian distance in four-dimensional joint space). This metric is only calculated and averaged for the successful cases. Besides, it is not computed for physically underactuated gripper because the joint angles during grasping are not accessible for the reasons mentioned above.
\end{itemize}

\begin{figure*}[t]
\centering
\includegraphics[width=1.0\textwidth]{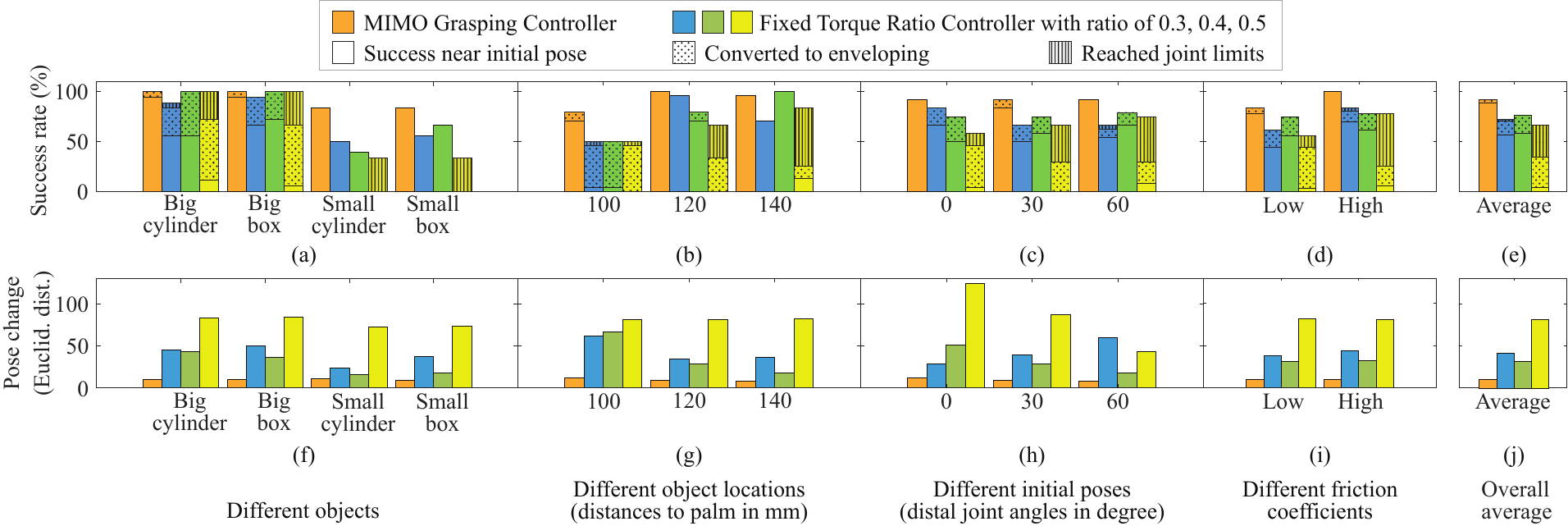}
\caption{Experiment results of fingertip grasping compared to Fixed Torque Ratio Controller.}
\label{fig:result}
\vspace{-3mm}
\end{figure*}

\subsubsection{Results}
Fig.~\ref{fig:pose} shows the photos of some typical scenarios in the experiments. (a) and (b) shows the resting pose and the initial touch. (c) shows the successful grasp near initial touch pose. The other images show cases in which the object was kept in fingertip grasp but a joint limit was reached (d), the grasp was transformed into an enveloping one (e), and the object was squeezed out of the hand (f). 

Fig.~\ref{fig:trq} shows that all controllers are effective in increasing joint torques. The horizontal axis shows different controllers (or grippers), the vertical axis is the magnitude of the four-dimensional torque vector ($\tau^{p1}_{meas}$, $\tau^{d1}_{meas}$, $\tau^{p2}_{meas}$, $\tau^{d2}_{meas}$) which indicates how ``strong'' the grasp is, and bar colors distinguish between initial touch and final grasp. We can see that there is a significant increase in the torque magnitude, and the torque levels in different cases are similar.

The results of the experiments comparing against Fixed Torque Ratio Controller are visualized as multiple bar charts in Fig.~\ref{fig:result}. In each plot, the bar colors show four different controllers,
the vertical axis is one of the performance metrics 
and the horizontal axis represents another dimension which is different in each plot (from (a) (f) to (d) (i): different objects, object locations, initial poses, and friction coefficients). In each bar in the first row showing the success rate, the pure-color area, the dotted area, and the line-shaded area represent, respectively, successful fingertip grasp without reaching joint limit, successful grasps but converted to enveloping, as well as successful fingertip grasp but joint angles reached limits.

\begin{figure*}[t]
\centering
\includegraphics[width=0.8 \textwidth]{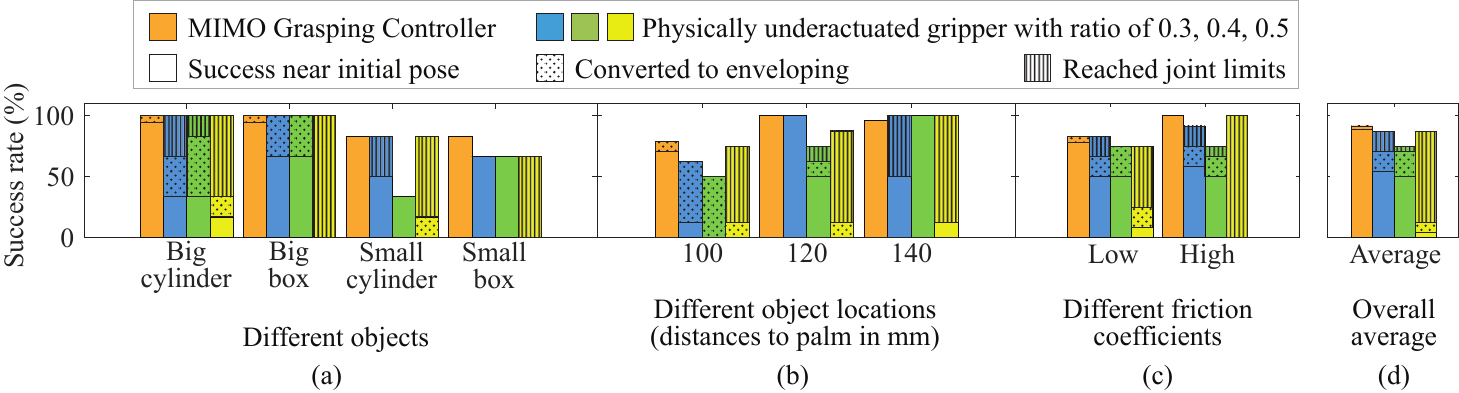}\\
\caption{Experiment results of fingertip grasping compared to physically underactuated gripper.}
\label{fig:result_ua}
\vspace{-2mm}
\end{figure*}

Similarly, the results comparing against physically underactuated grippers are shown in Fig.~\ref{fig:result_ua}. Here, the initial touch pose dimension is not available, so there are three dimensions (from (a) to (c): different objects, object locations, and friction coefficients). Also, the pose change metric is not available due to the absence of joint angle information. All other plotting rules are the same as Fig.~\ref{fig:result}.


As shown in Fig.~\ref{fig:result} (e)(j) and Fig.~\ref{fig:result_ua} (d), the overall success rates are 91.67\% for MIMO Grasping Controller, 72.22\%, 76.39\%, 66.67\% for Fixed Torque Ratio Controller with torque ratio of 0.3, 0.4, 0.5, respectively, and 87.50\%, 75.00\%, 87.50\% for physically underactuated gripper with torque ratio of 0.3, 0.4, 0.5, respectively. Even when the overall success rates are close (for example, Fig.~\ref{fig:result_ua} (d)), the types of the resulting grasps are significantly different. Besides, the gripper pose change metric for the MIMO controller is 9.96, compared to 41.55, 31.93, and 81.14 (degrees) respectively for the Fixed Torque Ratio controllers.

\subsection{Enveloping Grasps and Transitions}

We performed a second experiment to show that this gripper can perform enveloping grasp with either MIMO Grasping Controller or Fixed Torque Ratio Controller. We tested on two objects (big cylinder and big box), two object locations (60mm and 80mm from the palm), and two controllers mentioned above. The success rate is 100\%.

The last experiment is to show the performance of the transition from fingertip grasp to enveloping grasp. We first created fingertip grasps using the MIMO Grasping Controller, and then switched to Fixed Torque Ratio Controller with a ratio of 0.5. We tested on two objects (big cylinder and big box), three object locations (100, 120 and 140mm), three initial poses (0, 30 and 60 degrees) with the low friction fingertips. We found the success rate was 83.33\%.

\section{Discussion and Conclusion}

Overall, the results of the previous section support our hypotheses: the proprioceptive gripper running MIMO Grasping Controller is effective at executing stable fingertip grasps in a variety of situations and outperforms the baselines. Furthermore, the proprioceptive gripper exhibits versatility in being able to perform multiple types of grasps and also to transition between them on-demand.

Based on Fig.~\ref{fig:result} and \ref{fig:result_ua}, the MIMO Grasping Controller outperforms the baselines in fingertip grasping, and usually succeeds without transforming to an enveloping grasp or reaching joint limits. In contrast, the emulated and physical underactuated gripper often transform to an enveloping grasp or reach joint limits, thus relying on gripper reconfiguration. The second row of Fig.~\ref{fig:result} ((f) to (j)) gives similar intuition: for the Fixed Torque Ratio Controller, most cases have a large pose change. While the end-result is stable, it is different from the originally intended grasp. This might be unimportant or detrimental depending on the application.

It is also interesting to notice that, in different conditions, the optimal torque ratio for the emulated or physical underactuated gripper is different. We take this to mean that there is no one clearly preferable pre-set torque ratio, which could be physically implemented in a mechanical design, in order to obtain ideal performance in all these cases. In contrast, the proprioception-enabled gripper has the flexibility to alter torques at run-time.

Looking at how specific variables affect performance we can gain additional insights. From Fig.~\ref{fig:result} and~\ref{fig:result_ua} (a) and (b) we can see that the success rates for Fixed Torque Ratio Controller are low if the objects are small and close to the palm. This is because the emulated or physical underactuated gripper tends to transform the initial unstable fingertip grasps to enveloping when contacts are close to distal joints, but cannot cage the object if it is small because the distal links are fighting against each other --- a common issue for underactuated grippers. In contrast, the MIMO Grasping Controller does not suffer from this because it does not perform the conversion.


In the transitioning experiment, the high success rate shows the
proprioceptive gripper can indeed perform the conversion between grasp types
\textit{on-demand}. Though underactuated grippers also
occasionally perform such transitions, they occur unintentionally
and without giving the user an option to select the desired type of
grasp.

It is important to also highlight the limitations of this
study. Due to high dimensionality of the brute-force sweep in
our experiment, we cannot afford to cover a larger range with a finer
resolution for each dimension. In particular, we are unable to explore
more possibilities for physically implemented torque ratios. The
evaluation of the controller is primarily experimental and would
benefit from additional stability analysis, carried out for example
for representative cases and grasps.

Overall, we claim that proprioceptive manipulators, using active
sensing and control such as the MIMO Grasping Controller, represent a
promising way towards more versatile grasping and manipulation for
unknown objects. Future work will include the extension of the operation to three-dimensional cases, optimization / learning of the control gains, and the inclusion of hand
position to our set of actively controlled variables. We are aiming to
further explore these possibilities.

\addtolength{\textheight}{-12cm}   






\bibliographystyle{bib/IEEEtran}  
\bibliography{bib/sensing,bib/design,bib/control}  

\end{document}